\documentclass[sigconf]{acmart} 

\usepackage{graphicx} 
\usepackage[english]{babel} 
\usepackage{enumitem} 
\usepackage{titlecaps} 
\usepackage{enumitem}
\usepackage{hyperref}
\usepackage{booktabs}
\usepackage{siunitx}
\usepackage{makecell}
\acmConference[iWOAR]{iWOAR 2024 - 9th international Workshop on Sensor-Based Activity Recognition and Artificial Intelligence}{September 26--27,
  2024}{Potsdam, Germany}

\copyrightyear{2024}
\acmYear{2024}

\begin{document}

\title{Estimation of Psychosocial Work Environment Exposures Through Video Object Detection}
\subtitle{Proof of Concept Using CCTV Footage}

\author{Claus D. Hansen}
\email{clausdh@socsci.aau.dk}
\affiliation{%
 \institution{Department of Sociology and Social Work, Aalborg University}
 \city{Aalborg}
 \country{Denmark}
}
    
\author{Thuy Hai Le}
\email{thuyle21@student.aau.dk}
\affiliation{%
 \institution{Department of Computer Science, Aalborg University}
 \city{Aalborg}
 \country{Denmark}
}

\author{David Campos}
\email{dgcc@cs.aau.dk}
\affiliation{%
 \institution{Department of Computer Science, Aalborg University}
 \city{Aalborg}
 \country{Denmark}
}

\renewcommand{\shortauthors}{Hansen, Le \& Campos}

\begin{abstract}
    This paper examines the use of computer vision algorithms to estimate aspects of the psychosocial work environment 
    using CCTV footage. We present a proof of concept for a methodology that detects and tracks people in video footage
    and estimates interactions between customers and employees by estimating their poses and calculating the duration of
    their encounters. We propose a pipeline that combines existing object detection and tracking algorithms (YOLOv8 and
    DeepSORT) with pose estimation algorithms (BlazePose) to estimate the number of customers and employees in the footage
    as well as the duration of their encounters. We use a simple rule-based approach to classify the interactions as positive,
    neutral or negative based on three different criteria: distance, duration and pose. The proposed methodology is tested
    on a small dataset of CCTV footage. While the data is quite limited in particular with respect to the 
    quality of the footage, we have chosen this case as it represents a typical setting where the method could be applied.
    The results show that the object detection and tracking part of the pipeline has a reasonable performance on the dataset
    with a high degree of recall and reasonable accuracy. At this stage, the pose estimation is still limited to fully detect 
    the type of interactions due to difficulties in tracking employees in the footage. We conclude
    that the method is a promising alternative to self-reported measures of the psychosocial work environment and could be
    used in future studies to obtain external observations of the work environment.
\end{abstract}


\received{7 July 2024}

\maketitle

\section{Introduction}
    The use of Closed Circuit Television (CCTV) in public and private settings has increased substantially over the last 20 years.
    Its use is often associated with the prevention of crimes and aggressive behaviour, apprehension of criminals and raising
    the general perception of safety in public spaces. However, there are a number of other widespread uses of CCTV as well such
    as the estimation of traffic flows, automatic number plate recognition and the monitoring of wildlife and weather conditions. 
    Surveillance of people in shops and supermarkets are routinely used in order to prevent crimes and apprehend criminals
    that expose personnel for threats and assaults \cite{norris_growth_2004}. A meta-analysis of 76 studies on the use of CCTV
    in crime prevention showed that the use of CCTV in public places especially car parks reduced theft and other crimes 
    substantially.\cite{piza_cctv_2019} In a randomized controlled trial in the retail sector, Hayes \& Downs \cite{hayes_controlling_2011}
    showed that the use of surveillance in a retail setting reduced loss due to theft in shops with CCTV installed compared to the
    control group without CCTV. There are in other words some evidence for the usefulness of CCTV as a \textit{preventive measure}
    against crimes in general as well as theft in particular in the retail sector. 
    The use of surveillance is also justified by the usefulness of the data for \textit{investigating crimes} that have
    been committed. In a review from the UK, Ashby \cite{ashby_value_2017} showed that useful CCTV footage were available in 29\%
    of the cases of reported crimes on the British railway network. No such review has been carried out in the retail sector
    but we can assume that the footage would be even more useful here as the availability of the footage is
    higher. The use of CCTV in the retail sector is therefore justified both by the potential for preventing crimes and theft
    as well as the potential for investigating crimes that have been committed.

    As theft can sometimes lead to violence and threats against employees, the use of CCTV can also be seen as a preventive 
    measure against workplace violence and threats, creating a safer atmosphere for employees. This has been most explicitly 
    researched in the transport sector where CCTV as well as body cams has been used to prevent violence as well as other
    types of aggressive behaviour towards personnel.\cite{nobili_review_2023}  
    Fortunately, threats and violence are rare events in the retail sector. Arbejdstilsynet 
    (The Danish Working Environment Authority) bi-annually carries out surveillance on the work environment in Denmark
    through the survey NOA-L \cite{arbejdstilsynet_national_2021}. In 2021, 6.3 \% of the employees in the retail sector
    reported that they had been exposed to threats of violence in the work environment, a majority of these events were
    reported to be from external sources i.e. customers. A much smaller proportion of employees reported that they had been
    exposed to actual violent events at their workplace. 

    Because the rate of crime has been declining in most developed countries over a long period \cite{farrell_origins_2016},
    the consequence of the decline is that despite CCTV being very widespread most of the actual recorded footage is never
    used for anything at all. The question is whether this is a good or a bad thing?
     
    On the one hand, the use of CCTV footage is problematic even in the case of crime prevention because surveillance raises
    a number of privacy related questions. \cite{macnish_unblinking_2012} Michelman \cite{michelman_who_2009} argues
    that widespread surveillance of citizens in general can produce two kinds of harms: the first is the loss of privacy
    which is a basic right in modern democratic societies. Being able to move around freely without being caught on camera
    is a basic right that is threatened by the widespread use of CCTV. The loss of privacy is a harm in itself but it can also
    lead to other harms such as discrimination. Discrimination can occur when certain groups of people are
    singled out for surveillance because they are perceived to be more likely to commit crimes. This is a problem in particular
    when the surveillance is biased and the groups of people that are singled out are not actually more likely to commit crimes
    than other groups. The second harm that \cite{michelman_who_2009} identifies is called 'chilling effects' where 
    people are discouraged from engaging in certain activities because they fear that they are being watched. Criminals
    being deterred from committing crimes is a beneficial outcome of such 'chilling effects' of CCTV surveillance. 
    However, if the surveillance is perceived to be biased or unfair it might lead to certain groups of people refraining
    from going to the shop because they fear being singled out as potential criminals e.g. in the case where a theft is 
    committed in the shop and the police is called to investigate. For this reason, Macnish \cite{macnish_unblinking_2012}
    argues that a semi-automated (computational) approach to the processing of CCTV footage (e.g. object detection) is preferable
    to one that is solely based on human operators as it reduces the risk of human bias and discrimination. In addition, 
    such an approach is more efficient as the human operator can only process a limited amount of footage at a time. When 
    processing CCTV footage manually the limited 'processing capacity' of a human operator risks leading to the use of profiling
    and the influence of personal prejudices increases the risk of surveillance leading to stigmatisation and harassment of certain
    groups of people. Despite efforts to reduce negative effects of CCTV surveillance, there are
    still a number of ethical issues related to this and other forms of surveillance both in public and private settings
    which suggests that the use of CCTV should be limited as much as possible. In other words, the 'non-use' of CCTV footage
    is in itself a good thing because it reduces the risk of negative effects of surveillance.

    On the other hand, in the case where CCTV footage is being recorded and stored anyway, it could be argued that the
    'non-use' of the footage also carries problematic ethical issues~\cite{jones_other_2017}, such as the historical
    examples where companies or institutions decided to ignore information that was available to them because they did not
    want to act on it. We are reminded of this when it comes to e.g. harmful effects of fossil fuels, tobacco use and 
    asbestos where companies selling these products knew about the harmful effects but chose to ignore them in order to
    maximize profits~\cite{shearer_corporate_2015}. In the case of CCTV footage, the footage is being recorded and stored
    for the purpose of preventing in particular theft, i.e. primarily for the benefit of the company. However, the data
    could also be used for other purposes: it could for instance be used to advice customers on not to visit the shop at
    certain times of the day because the shop is too crowded which could be beneficial for the customers but also for the
    employees. Choosing not to use the data for these purposes could be seen as a missed opportunity to improve the work
    environment for the employees and the shopping experience for the customers. In this paper, we argue for the use of
    CCTV footage to estimate aspects of the psychosocial work environment and present a proof of concept demonstrating 
    how this can be achieved using a combination of existing object detection and computer vision algorithms.

    The psychosocial work environment is an important factor for employees health and well-being \cite{stansfeld_psychosocial_2006}.
    The global burden of work related accidents and diseases is substantial and is estimated to be 4\% of the global GDP
    and 2.7\% in Denmark \cite{takala_global_2014}. Although work accidents and occupational diseases related to e.g.
    exposures to harmful substances and physical work environment factors are more common, the psychosocial work environment
    is becoming an increasingly important factor for employees health and well-being as the labour markets of highly developed 
    countries such as Denmark are becoming more and more service oriented. For this reason, reducing adverse working
    conditions is important for society due to the costs associated with adverse work environment, e.g. prolonged sick leave and
    early retirement. At the same time a safe and healthy work environment is associated with increased productivity
    and reduced turnover which are important goals for companies. From the perspective of the individual, psychosocial work
    environment factors are important for a number of reasons not least because being outside the labour market has profound
    negative effects on the quality of life and longevity. Being able to estimate aspects of the psychosocial work environment
    using CCTV footage thus constitutes a way of using information that is already being collected for other
    purposes in a way that is beneficial for society and for the individual employees without incurring additional ethical
    problems that are not already present in the current limited use of CCTV footage for crime prevention. 
    
    The contribution of this paper is thus to provide a proof of concept that demonstrates how the use of object detection
    algorithms can be extended to estimate important aspects of the psychosocial work environment using CCTV footage. The
    contribution of this paper can thus be summarised as follows:
    \begin{itemize}
        \item Our paper proposes an implementation of object detection algorithms as a way of estimating quantitative job demands
        in the psychosocial work environment using CCTV footage.  
        \item We propose a set of metrics that can serve as proxy measures for important aspects of the psychosocial work environment.
        \item The methodology is tested on a dataset of CCTV footage establishing the feasibility of our
        approach.  
    \end{itemize} 

    \begin{figure}[h]
        \centering
        \includegraphics[width=0.49\textwidth]{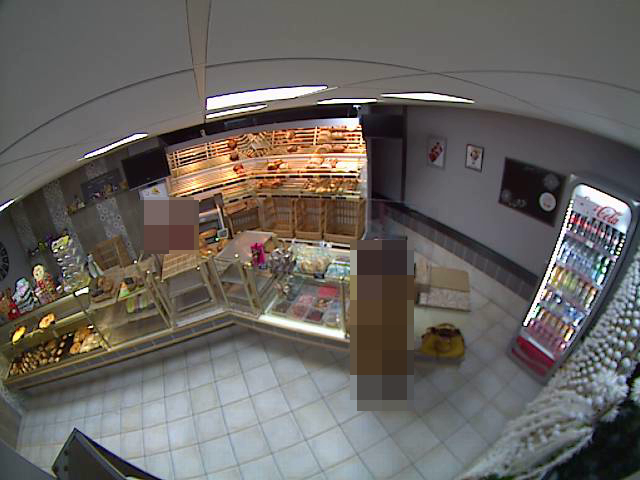}
        \caption{Example footage of an interaction between employee and customer in CCTV footage}
        \label{fig:bakery}
    \end{figure}

    We have structured the paper as follows: 
\begin{description}[leftmargin=*]
    \item[Section 2] reviews related work on computer vision algorithms for detecting and tracking people in video footage, 
    as well as recognising human activities using such footage. We also review how psychosocial work environment factors have
    been measured in previous studies and how the use of CCTV footage could provide a way of estimating some of these factors. 
    \item[Section 3] introduces the concepts and notation used in the paper. 
    \item[Section 4] describes the overall system architecture and methodology used to analyse the CCTV footage. The approach
    used combines several existing computer vision algorithms to detect and track people in the footage as well as to estimate
    the interactions between customers and employees. 
    \item[Section 5] presents the results of our experiments based on CCTV footage demonstrating the
    feasibility of our proposed method. 
    \item[Section 6] discusses the results and points to future work that needs to be carried out in order to validate the method
    further. We discuss the limitations of the method and how they can be addressed in future work.
\end{description}

\section{Related Work}

\subsection{Object detection, tracking and human activity recognition}

\subsubsection{Object detection}
\textit{Object detection} in images and video footage is a core task in computer vision \cite{zou_object_2023}. Early approaches,
like Histogram of Oriented Gradients (HOG) \cite{dalal_histograms_2005} and Haar cascades \cite{viola_rapid_2001}, relied on 
'hand-crafted' features. Modern methods, however, use deep learning to automatically learn features from data. One of the most
influential algorithms is You Only Look Once (YOLO) \cite{redmon_you_2016}, which detects objects in real-time by dividing the
image into a grid and predicting bounding boxes and class probabilities within each cell. Although YOLO is not the most precise,
its speed makes it suitable for real-time applications or applications running on devices with limited computational resources. 

In CCTV footage analysis, object detection helps identifying harmful objects and individuals. \cite{salazar_gonzalez_real-time_2020}
Challenges include detecting small objects and dealing with low-quality footage, which might require fine-tuning or training
the models on domain-specific data. \cite{zou_object_2023} 

\subsubsection{Multi-object tracking}
\textit{Multi-object tracking} (MOT) extends object detection by tracking objects across frames.\cite{luo_multiple_2021} Approaches vary between online and 
offline tracking, with online tracking being crucial for real-time applications such as autonomous vehicles and CCTV footage
analysis.\cite{agarwal_abandoned_2018} One prominent algorithm is Simple Online and Realtime Tracking (SORT) \cite{bewley_simple_2016},
which uses a Kalman filter to predict object trajectories based on bounding box movements. SORT is efficient but struggles with
occlusions and non-linear movements. DeepSORT \cite{wojke_simple_2017} enhances SORT by integrating deep learning for 
appearance-based feature extraction, significantly improving tracking accuracy by reducing identity switches.

In crowded environments like shops, precise tracking of individuals is critical.\cite{sivakumar_real_2021} Combining object detection
with tracking algorithms helps estimate core aspects of the psychosocial work environment, such as the number of customers and 
employees and the duration of their interactions. Object tracking however is not sufficient for analysing interactions
taking place between customers and employees in the footage. For this purpose we need to be able to assess the nature of the
activities that the people are carrying out in the footage which can be done using human activity recognition algorithms.

\subsubsection{Human activity recognition}
One type of \textit{human activity recognition} (HAR) involves identifying actions from video footage.\cite{arshad_human_2022} Deep learning
algorithms like CNNs are commonly used, with applications focusing on daily activities rather than workplace settings. Notable HAR
algorithms include OpenPose \cite{cao_openpose_2021}, which estimates poses with up to 135 key-points using Part Affinity Fields,
and MediaPipe Pose \cite{noauthor_mediapipe_nodate}, based on BlazePose \cite{bazarevsky_blazepose_2020}, which uses 33 key-points
and is optimized for real-time mobile applications.

Examples of HAR applications include Ruiz-Santaquiteria et al. \cite{ruiz-santaquiteria_handgun_2021}, which combines object 
detection and pose estimation to identify handguns, and Paudel et al. \cite{paudel_deep-learning_2020}, which estimates the
ergonomic risk of workers' poses. These applications demonstrate the potential of pose estimation for recognizing specific
human activities in video footage including the identification of interactions related to the psychosocial work environment.

Despite the availability of algorithms for object detection and tracking, no existing solution address the specific task of
estimating psychosocial work environment factors from CCTV footage. Our approach combines object detection, tracking, and pose
estimation to recognize and classify interactions between customers and employees, providing a novel method for estimating
psychosocial work environment factors.

\subsection{Workload estimation and psychosocial work environment factors}

\subsubsection{Self-reported estimation of psychosocial work environment factors}
The job demands-control model by Karasek \cite{karasek_jr_job_1979} is widely used to estimate the psychosocial work
environment, focusing on job demands and job control. High job demands are linked to negative health outcomes like
cardiovascular disease and depression \cite{stansfeld_psychosocial_2006}. Most research relies on self-reported data,
making it vulnerable to biases and subjective evalutations \cite{nieuwenhuijsen_psychosocial_2010}. Theorell and Hasselhorn
\cite{theorell_cross-sectional_2005} suggest supplementing self-reports with objective measures like expert ratings or 
job exposure matrices. Our paper contributes by proposing a computational approach to analyse CCTV footage, offering an
alternative to traditional self-reported measures.

\subsubsection{Observational estimation of psychosocial work environment factors}
Few studies use person-independent measures for psychosocial work environment factors. Waldenstrom et al. 
\cite{waldenstrom_externally_2008} used expert ratings for social support, Bosma et al. \cite{bosma_job_1998} 
used expert ratings to estimate both levels of skill discretion as well as job control, while Griffin et. al. 
\cite{griffin_effect_2007} used external observers to rate skill utilization and job barriers. These methods are costly
and time-consuming, limiting their scalability. Imputed approaches estimate job demands based on average levels in
occupational groups, but face validity issues, especially for small groups \cite{niedhammer_study_2018}. 

Greiner et al. \cite{greiner_occupational_2004} used observational data to estimate work barriers and time pressure, showing
a link to blood pressure. However, this method is not scalable for larger populations. Routine CCTV monitoring in shops
offer a unique opportunity to obtain external observations of the psychosocial work environment factors. Using CCTV footage
is cost-effective and scalable, unlike expert ratings. 

\begin{figure}[h]
    \centering
    \includegraphics[width=0.49\textwidth]{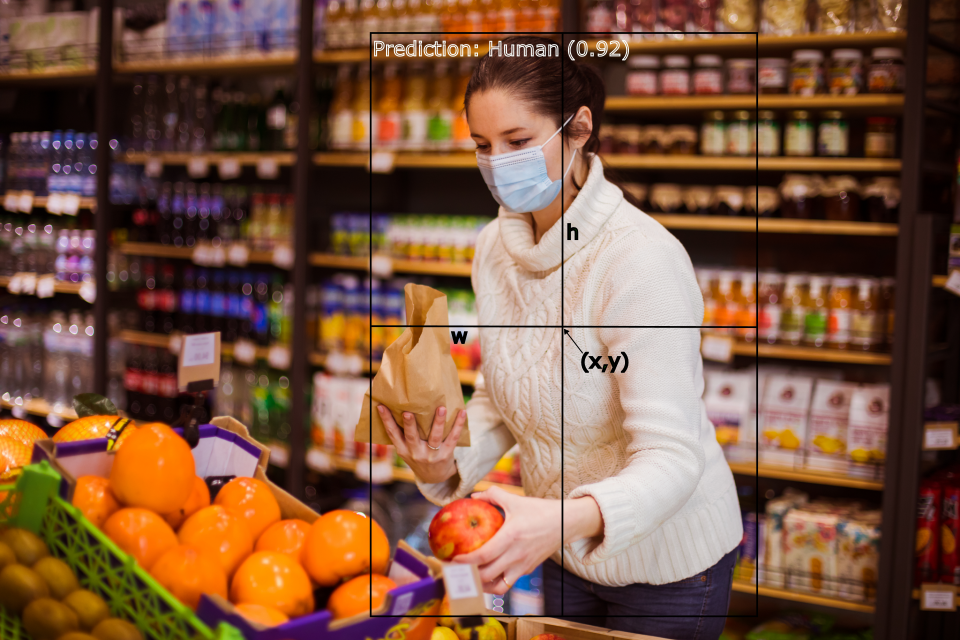}
    \caption{Bounding box example with predictions $(x, y, w, h)$ and class label. Photo: Colourbox.com}
    \label{fig:bbox}
\end{figure}

\section{Concepts and notation}
In the following, we will introduce some notation and concepts that form the foundation of the methodology used in this paper.

\subsection{Object-detection and multi-object tracking using bounding boxes}
Bounding boxes are a common way of representing objects in images and video footage (see Figure \ref{fig:bbox}). A bounding box
is a rectangle that encloses an object in the image or frame. The bounding box is defined by a tuple of four numbers 
$(x, y, w, h)$ where $x$ and $y$ are the coordinates of the center of the bounding box and $w$ and $h$ are 
the width and height of the bounding box respectively. The bounding box is used to represent the position of the object
detected in the frame and is used as input to algorithms that subsequently track the object across frames such as DeepSort.
Bounding box regression involves predicting the boundaries of the rectangle that encloses an object in the image as well
as the class of the object and the confidence with which the predicting has been made. In Figure \ref{fig:bbox}, the bounding
box is used to represent the current position of the object in the frame as well as the predicted class of the object and the
confidence with which the prediction has been made. 

Multi-object tracking is a technique for tracking multiple objects in video footage. The technique is based on object detection
algorithms that are able to detect objects in individual frames and track the objects across frames. The tracking is based on
the movement of the bounding boxes of the objects in the frames and is used to estimate the trajectories of the objects in the
video footage.

\subsection{Human activity recognition using landmark pose detection}
\begin{figure}[!h]
    \centering
    \includegraphics[width=0.33\textwidth, height=0.33\textheight]{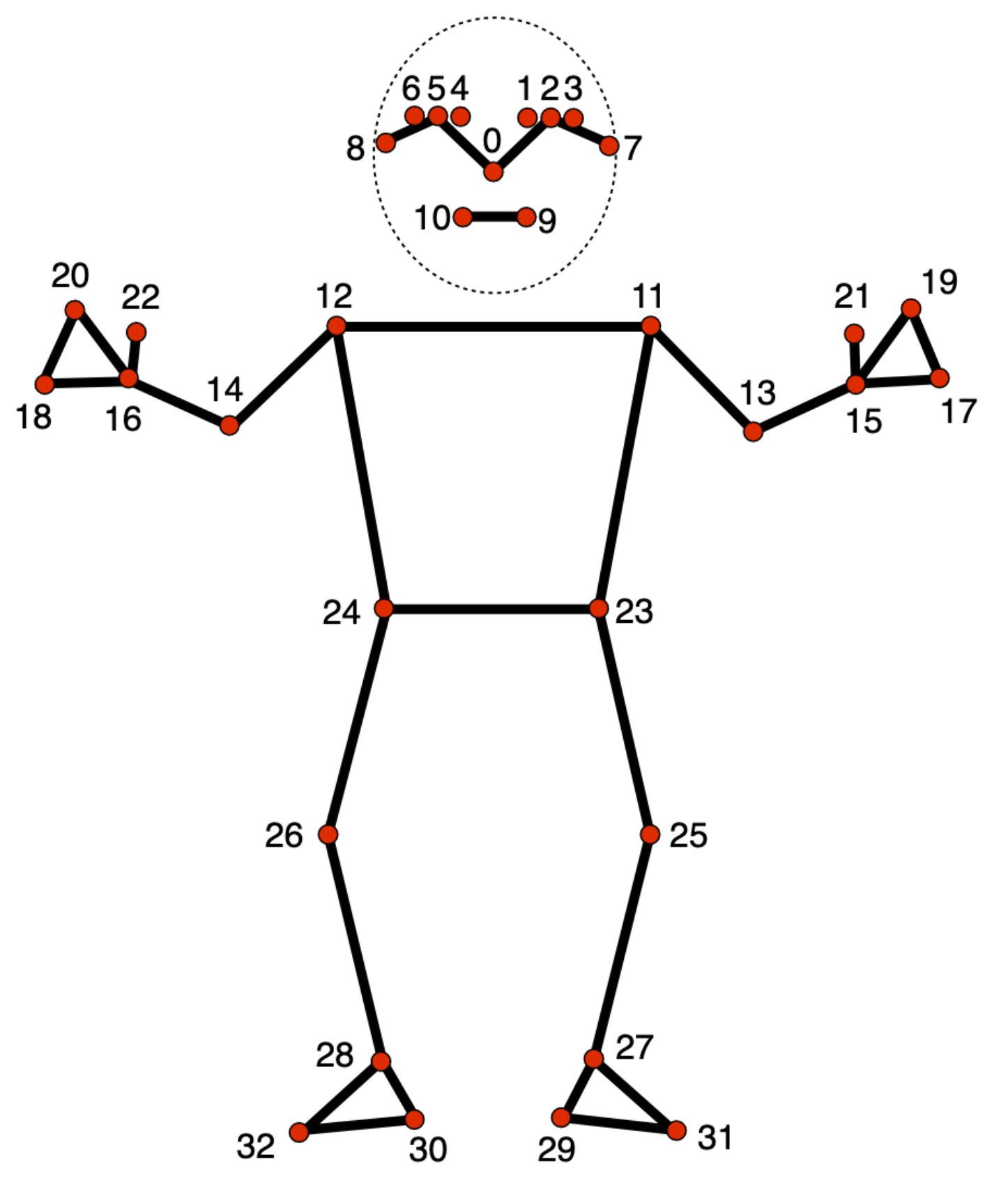}
    \caption{Pose estimation example with 33 key-points}
    \label{fig:pose}
\end{figure}

Landmark pose detection is a technique for estimating the pose of a person in an image. Depending on the level of detail the 
pose is defined by a number of key-points that are placed on the person's body and refers to specific parts of the body such as
head, shoulders, knee, hands etc. The BlazePose algorithm uses 33 key-points to estimate the pose of a person in an image as 
illustrated in Figure \ref{fig:pose}. The key-points are used to estimate the position of the person's body parts
and are used as input to human activity recognition algorithms that are able to estimate the activities that the person is
carrying out in the image or frame.

\subsection{Proxy metric of quantitative job demands}
Job demands are defined as the psychological load that the employee experiences on a normal work day. This construct is not
one-dimensional but consists of a number of different types of demands such as quantitative demands (e.g. number of work hours
and pace of work), emotional demands (e.g. dealing with difficult customers) and cognitive demands (e.g. problem solving and
decision making) \cite{kristensen_copenhagen_2005}. Using a computational approach to estimate job demands from CCTV footage
does not lay claim to be able to estimate all of these aspects of job demands but focuses on a subset of what could be 
considered elements of quantitative demands \cite{kristensen_distinction_2004}.

Our aim is to estimate the tuple $V_{\text{job demands}}$ that represents the quantitative job demands experienced by the
employees in the shop.

The tuple \( V_{\text{job demands}} \) is defined as:
\[
V_{\text{job demands}} = (T, C, S, D_{\text{total}}, D_{\text{avg}})
\]
where:
\begin{itemize}
  \item \( T \) total duration of work day (in minutes),
  \item \( S \) Set of all encounters \( \{E_1, E_2, \ldots, E_n\} \),
  \item \( C \) Total number of customers, calculated as the sum of customers in each encounter: \( C = \sum_{E_i \in S} \text{n}_i \).
  \item \( D_{\text{total}} \) total duration of the encounters (in minutes), calculated as \(\sum_{E_i \in S} \text{duration}_i\).
  \item \( D_{\text{avg}} \) average duration of the encounters (in minutes), computed as \(\frac{1}{|S|} \sum_{E_i \in S} \text{duration}_i\).
\end{itemize}

Each encounter \(E_i\) within the set \(S\) is defined as:
\[
E_i = (\text{id}_i, \text{duration}_i, \text{class}_i, \text{count}_i)
\]
where:
\begin{itemize}
  \item \( \text{id}_i \) is the unique identifier of the encounter,
  \item \( \text{dur}_i \) is the duration of the encounter (in minutes),
  \item \( \text{class}_i \) is the class of the encounter (positive, neutral, negative),
  \item \( \text{n}_i \) is the number of customers in the encounter.
\end{itemize}

Looking more specifically at the nature of the interactions between the customers and employees we can make a basic distinction
between negative encounters (e.g. involving threats and violence in the worst case), neutral encounters (e.g. involving regular
customer service with a neutral tone and blasé attitude from both parties) and positive encounters (e.g. involving a friendly
tone and possibly small talk between the customer and the employee). We detail in the methodology section how these distinctions
can be calculated using a combination of pose estimation and a simple rule-based approach based on the duration of the encounters
and the distance between the bounding boxes of the customers and the employees.

\section{Methodology}

\begin{figure*}[t]
    \centering
    \includegraphics[width=\textwidth]{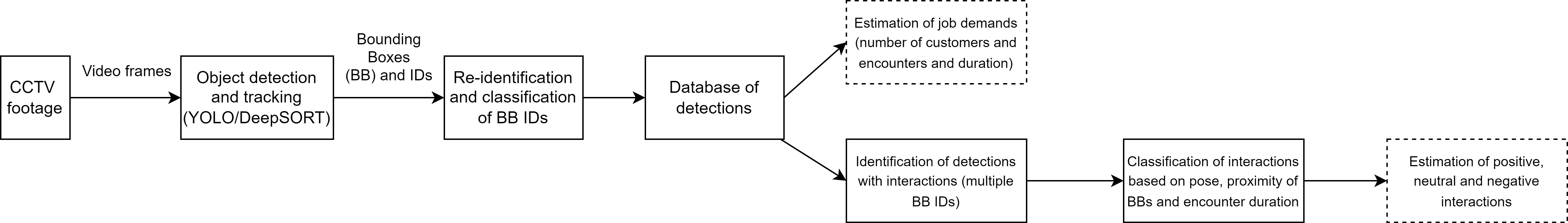}
    \caption{Overview of system architecture}
    \label{fig:seccamarchitecture}
\end{figure*}

\subsection{System architecture}

In Figure \ref{fig:seccamarchitecture}, we present an overview of the system architecture that we use to estimate job demands
from CCTV footage. As can be seen from the illustration, the system consists of a fundamental pipeline that detects and tracks
persons in the video footage classifying them into customers and employees respectively. This classification is stored in a
temporary database that is subsequently used to estimate job demands based on the number of customers, the number of encounters
and their duration. Building on this classification, frames with at least two bounding boxes are selected for further analysis
using the pose estimation algorithm as well as a simple rule-based approach (as explained below) to classify the interactions
into positive, neutral or negative encounters. This enables us to estimate the final aspects of job demands arising from the
nature of the interactions.

\subsection{Object detection and tracking with YOLOv8 and DeepSORT}
This section details the system employed for object detection and tracking in the video footage. We make use of a combination
of two widely used deep learning algorithms: YOLOv8 \cite{jocher_ultralytics_2023} for object detection and DeepSORT 
\cite{wojke_simple_2017} for object tracking. Unlike traditional methods that 
require separate stages for feature extraction and classification, YOLOv8 operates on a single-stage approach. This means
it divides the input image into a grid and simultaneously predicts bounding boxes and class probabilities for objects contained
within each grid cell. We restrict the object detection to be persons only, as we are only interested in being able to distinguish
between customers and employees in the footage. The unique approach of YOLOv8 allows 
it to achieve real-time object detection speeds without compromising on detection accuracy. Although our system is not designed
to be used in real-time, the efficiency of YOLOv8 is beneficial for processing large amounts of video footage. This would also
be a requirement if the system should be set up to run on the hardware on-site used for recording the CCTV footage (i.e. some kind
of webcam with limited computational resources). 

The detection from YOLOv8 (in the form of bounding boxes, class predictions and confidence scores) is saved in a temporary database
and subsequently used as input to the DeepSORT algorithm. DeepSORT utilizes these detections and extracts features from the objects
to associate these across frames, assigning unique IDs and maintaining track information for each object throughout the video. 
We use the pre-trained model based on the MARS dataset \cite{zheng_mars_2016} which is a large-scale dataset for person re-identification
trained on a large number of pedestrian images. This provided us with a quick and efficient way to track persons across frames 
without the need to train the model on our own dataset. Because of the basic assumptions of the DeepSORT algorithm, however, this
can lead to errors in tracking objects across frames. The Kalman filter used as the basis for DeepSORT assumes that the objects 
move in a linear fashion and that the velocity of the objects is approximately constant \cite{welch_introduction_1995}. 
While this is a reasonable assumption for pedestrians walking around in public spaces, this is a more problematic assumption 
in the context of the retail sector. For this reason, DeepSORT IDs be a suboptimal approach. In some cases, the same object might
be assigned to a different ID, especially in situations with occlusions, similar-looking objects, changes in the shape of the bounding
box associated with an object or with movement of persons that are not linear. This is particularly the case in our footage in situations
where the employee bow downs to pick up something or when the employee serves the customers by reaching out to the shelves. This will
lead to errors when calculating properties or tracking specific objects based solely on their DeepSORT IDs.

\subsection{Re-identification of bounding box IDs and classification of customers and employees}
\begin{figure}[h]
    \centering
    \includegraphics[width=0.49\textwidth]{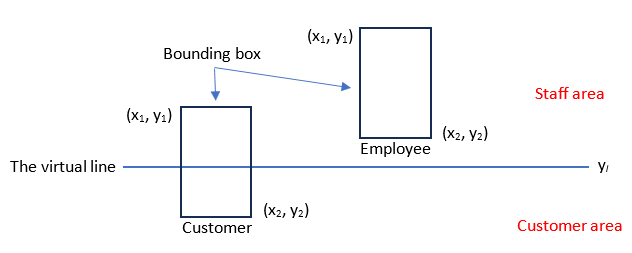}
    \caption{Line $l$ dividing the frame into two regions of interest (ROI) for classification of customers and employees}
    \label{fig:roi}
\end{figure}

Because the object detection algorithm is only able to detect persons regardless of their role as customers or employees, we need
to implement a method to distinguish between the two. Even if we had trained our algorithm on a dataset from the retail sector,
it would be unlikely that it would be able to distinguish between the two classes of persons based solely on their appearance. For
this reason, we have implemented a simple rule-based approach that classifies the persons based on their initial position in the
frame. As can be seen from the illustration in Figure \ref{fig:roi}, we have defined two regions of interest (ROI) in the frame 
based on the line $l$ that divides the frame into a \textit{customer area} where the entrance to the shop is located and a
\textit{staff area} where the counter is located and the employee can exit to a back room. The classification of objects is
based on the y-coordinate of the bottom-right corner of the bounding box $y_2$: bounding boxes with $y_2 < y_l$ (i.e. less than
the y-coordinate of $l$) are classified as \textit{employees} while bounding boxes with $y_2 > y_l$ are classified as 
\textit{customers}. This simple rule helps reduce the computational load for re-identifying IDs for employees, as all
IDs of bounding boxes in the staff area is reassigned to the employee. This is build on an assumption that there is only
one employee in the staff area at a time. In more complex settings, the definition of the ROI could be extended to include
multiple employees and a more sophisticated division of the frame into different areas. 

In an attempt to supplement the tracking IDs derived from DeepSORT, we have implemented a combination of three additional
methods that analyse features within the bounding boxes identified by YOLO. These methods include Convolutional Neural Networks
(CNNs) (and more specifically the EfficientNet-B0 model \cite{tan_efficientnet_2020}) for feature extraction, Histogram of
Oriented Gradients (HOG) for shape analysis, and K-Means clustering for color analysis. If the similarity score between two
bounding boxes based on the features extracted by the three methods is greater
than a predefined threshold, the program re-identifies the objects and assigns them the same ID, ensuring consistent tracking.
By combining the three methods, we aim to achieve more robust and accurate results compared to using the individual methods
in isolation. Our solution is inspired by Ye and colleagues \cite{ye_deep_2021} who reviewed state-of-the-art methods for
re-identifying objects in video footage and points to the importance of combining multiple methods to achieve the best results.
Our solution is designed to be able to weight the contribution of each of the three re-identification methods to the total
similarity score.

The process for re-identifying IDs for objects is performed following these steps:
\begin{itemize}
    \item Scan all bounding boxes to calculate CNN, HOG, and K-Means features for each ID assigned by DeepSORT.
    \item For each pair of IDs (\(Object_1, Object_2\)), calculate the similarity between them based on their CNN, 
    HOG, color (K-Means) features.
    \item Combine the similarities using a weighted average where the weights are chosen based on a predefined importance 
    of the features in distinguishing between objects.
    \item Filter pair IDs that are greater than the threshold and sort them in descending order of similarities.
    \item Re-identify IDs by adding all bounding boxes of \(Object_2\) to \(Object_1\) and deleting \(Object_2\).
\end{itemize}

This process is repeated for all pairs of IDs and is designed to be able to re-identify objects that e.g. have been occluded, 
whose bounding box have changed shape in the frame or have been wrongly assigned a new ID by DeepSORT. The weights for the CNN,
HOG and color similarities are chosen based on the importance of the features in distinguishing between objects. In our setting,
we priorize high-level features due to the relatively low level of detail in the footage which means attributing most weight ($0.65$)
to the EfficientNet-B0 model as it is able to capture high-level features crucial for distinguishing
between different objects based on their overall structure and appearance. The HOG features are given a moderate weight $(0.20)$ as
they capture shape and edge information that is useful but often less distinctive than high-level features from CNNs. Finally, the
color features could help differentiate between customers and employees based on the color of their clothing and are given the
remaining weight $(0.15)$.

The re-identified IDs are used to estimate the number of customers entering the shop as well as the duration of their encounters.
The duration of the encounters is calculated by taking the number of bounding boxes assigned to the customer and dividing it by
the frame rate of the video footage. This approach assumes that the encounters begin when the customer enters the shop and ends
when the customer leaves the shop. Using the pose-estimation approach we base the calculation of the duration of the encounters
on the proximity of the bounding boxes of the customers and the employees. When the distance between the bounding boxes is less
than a predefined threshold (distance $< 1.5 \, \text{m}$), the encounter is considered to have started. 

\subsection{Human activity recognition}

\begin{figure}[h]
    \centering
    \includegraphics[width=0.49\textwidth]{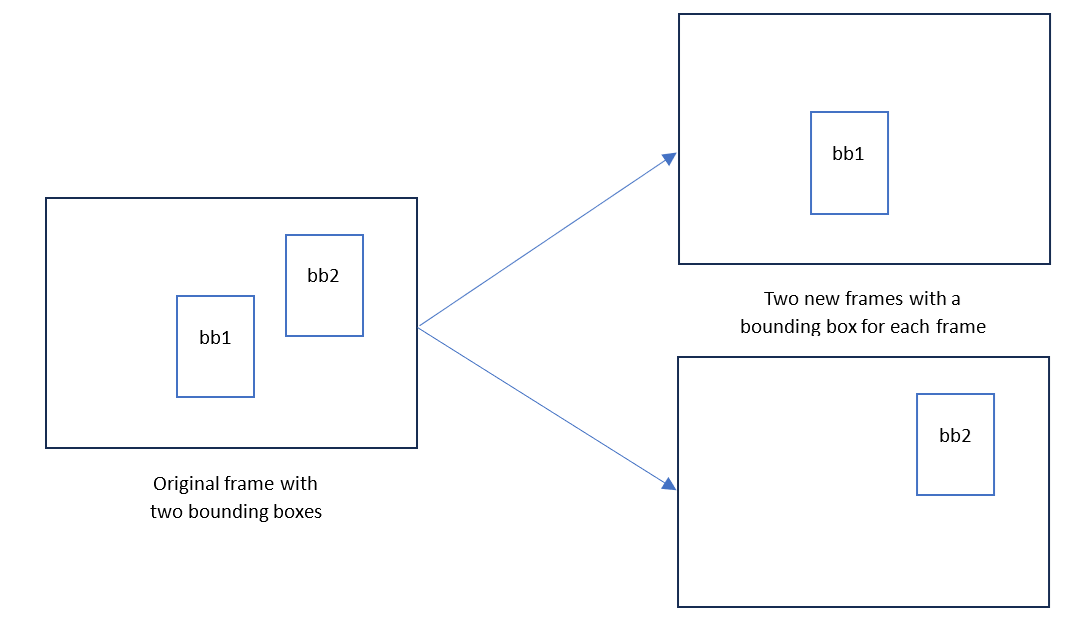}
    \caption{Splitting the frame into multiple frames with isolated bounding boxes for pose estimation}
    \label{fig:bbsplit}
\end{figure}

As can be seen from the illustration in Figure \ref{fig:seccamarchitecture}, the results from the object detection and tracking
are subsequently used to select frames with at least two bounding boxes for further analysis, i.e to identify \textit{interactions}
between customers and employees. We utilize \textit{MediaPipe Pose solution} to estimate
the landmark poses of the customers and employees in the selected frames. The algorithm is based on a pre-trained model that predict
33 keypoints of a person's skeleton, allowing us to track the movements and interactions of the individuals in the frame. 
While MediaPipe Pose excels at pose estimation, it is designed to process a single person per frame only. To overcome this limitation
when dealing with multiple people in a video, a potential approach involves strategically manipulating the frames as illustrated
in Figure \ref{fig:bbsplit}. The procedure is thus as follows:
\begin{enumerate}[label=\alph*.]
\item For each frame in the original video containing multiple people a new empty frame with 
the same dimensions and data type is created.
\item Loop through each bounding box detected by YOLOv8 in the original frame.
\item Copy the corresponding bounding box information (coordinates, width, height) from the
original frame to the empty frame.
\item	Apply the MediaPipe Pose algorithm to this newly created frame containing only the isolated bounding 
box to accurately estimate that person's key-point coordinates.
\item Repeat steps a-d for each detected bounding box in the original frame.
\end{enumerate}

On the basis of the landmark poses obtained as well as information about the placement of the bounding boxes in the frame
relative to each other, we then apply another rule-based approach to classify the interactions between customers and employees.
The aim is to be able to distinguish between positive, neutral and negative interactions based on the combined information.
As desribed above the distinction between the three types of interactions is based on three different criteria: \textit{duration}
of the encounter, \textit{distance} between the bounding boxes of the two persons and their placement with respect to staff and
customer area and finally, the \textit{pose of the persons} in the frame.

\subsubsection{Neutral interactions}
We assume that neutral interactions are the most common type of interaction in the retail sector and that overtly negative
interactions are the least common. A neutral interaction is defined as a scenario where an employee situated in the staff
area carries out tasks related to the servicing of customers such as retrieving bread from the shelves, packing goods
and registering the payment at the counter. We assume that a neutral interaction of the encounter $E_i$ is characterized 
by a duration of the encounter that is close to the average duration of all encounters or shorter than the average duration
of all encounters (i.e. \(dur_i \lesssim D_{\text{avg}}\)), and that the distance between
the two persons equals the distance between a customer near the counter and the employee at the opposite side of the counter. 

\subsubsection{Positive interactions}
We assume that positive interactions are characterized by a friendly tone and possibly small talk between the customer and
the employee. This implies that the duration of the encounter is longer than the average duration of all encounters (i.e.
\(dur_i > D_{\text{avg}}\)) and either 
\begin{itemize}
\item that the employee is standing still at the counter without performing any tasks related to
the servicing of customers or,
\item that the employee is coming out from the staff area to the customer area. 
\end{itemize}

We use the position of the bounding boxes to determine the placement of the persons in the frame in order to distinguish
between the two scenarios.

\subsubsection{Negative interactions}
Although negative interactions could be characterized by a number of different behaviours, we focus on one specific type of
negative interaction in this study: threatening or violent behaviour. We assume that threatening behaviour is characterized
by a customer trying to intimidate the employee by raising their hands above their head or hitting the employee. This means 
that the duration of the encounter is not relevant for the classification of the interaction. We use the landmark poses 
obtained from the Mediapipe algorithm. 

\begin{figure}[h]
    \centering
    \includegraphics[width=0.49\textwidth]{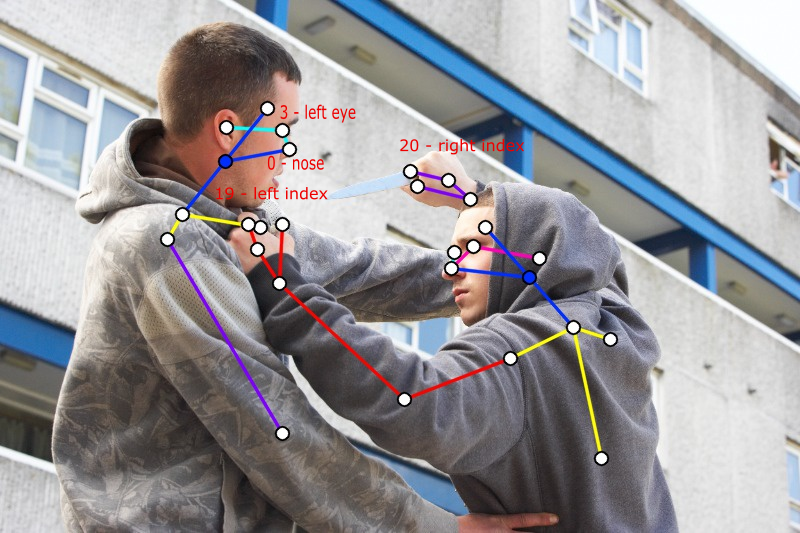}
    \caption{Example of threatening pose with keypoint notations $(0, 3, 19, 20)$. Photo: Colourbox.com}
    \label{fig:threatpose}
\end{figure}

As shown in Figure \ref{fig:threatpose}, we can calculate the distance between the hands (19, 20) of one object and the
head (0, 3) of another object within the same frame. An interaction is classified as negative if the calculated distance
is less than the threshold, which typically ranges from 0 to 5 cm. This threshold is based on the assumption that threatening
behaviour is characterized by a close proximity between the hands of the customer and the head of the employee. 

It would be possible (and advantageous) to include more types of poses that are threatening e.g. a customer pointing a finger
at the employee or poses signalling that the customer is angry or frustrated. This would require a more detailed analysis of
the poses obtained from the Mediapipe algorithm and the implementation of a more complex rule-based approach to classify the
interactions. As our study is a proof of concept we have chosen to only focus on this specific type of threatening behaviour
despite the fact that it is highly unlikely for this to be prevalent in the video footage we analyse. 

\subsection{Estimation of psychosocial work environment exposures}
Based on the classification of customers and employees, we are able to estimate $V_{\text{job demands}}$ 
as described in the previous section. The total number of customers \(C\) is calculated as the sum of customers in each encounter
and the total duration of the encounters \(D_{\text{total}}\) is calculated as the sum of the duration of each encounter. The
average duration of the encounters \(D_{\text{avg}}\) is calculated as the average duration of all encounters. The duration of
the encounters is calculated based on the frame rate of the footage and the number of frames in each encounter.

\section{Experiments and Results}
We use video footage captures from shops and malls. 
The footage is of low quality and the number of persons in the frame
is quite limited, however, we believe that the material is representative of the type of footage that could be expected from
CCTV cameras in the retail sector. As the aim of our paper is to provide a proof of concept that the method could be used to
estimate aspects of the psychosocial work environment using CCTV footage, we find this fitting. We utilize a total of 1500
minutes of footage from a bakery and have analysed a subset of this footage consisting of 5 videos totalling 73 minutes.
In the following, we describe the results of the experiments carried out as part of our study
including information about the setup used.

\subsection{Setup}
The tests carried out as part of this study were all run on a laptop with an Intel Core i7-10750H CPU and 16 GB of RAM. The
laptop was running Windows 11. We have made use of different existing libraries and have written our own code to implement
the methodology described in the previous sections using Python 3.11.  

\subsection{CCTV footage analysis}
The CCTV footage was captured as MJPG which means that the frame rate is not constant across the footage 
\cite{noauthor_motion_2024}. The frame rate most likely varies between 10-15 frames per second. For calculating the duration
of the encounters, we have assumed a frame rate of 15 frames per second. The resolution of the footage
is 640x480 pixels. The camera is placed in the ceiling of the workplace with a view of the counter and the entrance to the
shop however without any information about the distance between the camera and the persons in the frame or the angle of
the camera. These information could have helped improve the accuracy of the different algorithms used in the study for 
instance when estimating the distance between the bounding boxes of the persons in the frame. Depending
on the way in which a customer enters the shop they might not be detected by the object detection algorithm as they are 
partly occluded. 

\subsection{Coding of video footage}
In order to evaluate the performance of our methodology, we have coded the five videos by watching them and manually coding
how many customers come into the shop and for how long they stay. In addition, we have coded the encounter as either 
positive, neutral or negative based primarily on the criteria described in the methodology section: positive encounters
are characterised by the employee not performing any tasks related to the servicing of customers but seems to be talking
to them and the duration of the encounter is therefore longer than the average neutral encounter. Neutral encounters are
coded as such when the customer comes into the shop, buys an item and then exits
the shop again immediately thereafter. We have not coded any negative encounters as the footage does not contain any 
threatening behaviour. 

We decided not to annotate the bounding boxes of the persons in the frame as this 
is a time consuming task that would primarily be related to the accuracy of the object detection
and tracking algorithm which is not the primary contribution of this study. The coding of aggregate measures related to the
videos enables us to compare the results of the methodology 
with the manually coded data and calculate the accuracy of our proposed method with respect to three elements from 
\(V_{\text{job demands}}\), namely the total number of customers \(C\) and the total \(D_{\text{total}}\) as well as 
average duration of the encounters \(D_{\text{avg}}\). The accuracy of the predictions is calculated as the percentage of
correctly identified customers and encounters as well as the average difference between the predicted and the manually
coded values for the total and average duration of the encounters. These metrics are evaluated for each of the five videos
and the average accuracy is calculated across the five videos. The results of these calculations are presented in table 
\ref{table:objdet}.

\begin{table*}[h]
    \centering
    \caption{Accuracy of object detection and tracking algorithm}
    \label{table:objdet}
    \renewcommand{\arraystretch}{1.5}
    \begin{tabular}{ccccccc}
    \toprule
    Video Nr. & \thead{No. of\\ customers} & \thead{Accuracy\\ (No. customers)} & \thead{Recall\\ (No. customers)} & \thead{Average duration\\ (min:sec)} & \thead{Error\\ (Duration)} & \thead{Ratio of Positive\\ interactions} \\
    \midrule
    1 & 6 & 86\% & 100\% & 1:23 & 30\% & 12\% \\
    2 & 14 & 43\% & 43\% & 1:54 & 33\% & 28\% \\
    3 & 1 & 100\% & 100\% & 1:09 & 7\% & 0\% \\
    4 & 4 & 80\% & 100\% & 1:05 & 17\% & 0\% \\
    5 & 4 & 31\% & 100\% & 3:46 & 58\% & 83\% \\
    Average &  & 68\% & 89\% &  & 29\% &  \\
    \bottomrule
    \end{tabular}
\end{table*}

The results of the experiments carried out as part of this study show that the methodology is able to roughly estimate $C$, the 
number of customers in the footage with an accuracy of 68\% and a recall of 89\%. The approach, however, underestimates 
\(D_{\text{average}}\), the average duration of the encounters by on average 29\%.

\subsection{Object detection and tracking}
\begin{figure}[h]
    \centering
    \includegraphics[width=0.49\textwidth]{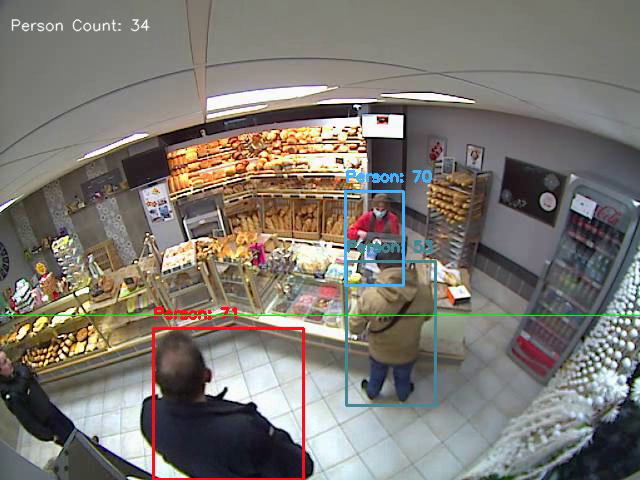}
    \caption{Example of object detection and tracking with bounding boxes in the CCTV footage}
    \label{fig:objdetexample}
\end{figure}

As can be seen from Figure \ref{fig:objdetexample}, the object detection and tracking algorithm is able to detect and track
persons in the frame and applying our rule-based approach classify the persons as customers and employees respectively. 
The algorithm struggles primarily with re-identifying the employee across frames. This is likely due to the limitations of
the DeepSORT algorithm as described in the methodology section, i.e. the assumption that the objects move in a linear fashion
and that the velocity of the objects is approximately constant. The difficulties in re-identifying the employee across frames
has consequences for the analyses of interactions as the employee is not detected in the frames in a high number of cases. 

As can be seen in Table \ref{table:objdet} the accuracy and precision of the object detection and tracking algorithm is best
in videos with a low number of customers characterised by a neutral interaction pattern where the customer enters the shop, order
an item, pays for it and then exists the shop immediately thereafter. In these cases (i.e. videos 3 and 4), the algorithm
is able to detect and track the customers with a high degree of recall and accuracy. The worst performance is seen in video 5
where only 4 customers visit the shop (in the span of 15 minutes) and where one positive encounter takes a very long time. The
algorithm detects this particular customer multiple times leading to an overestimation of the total number of customers as well as
severely underestimating the duration of the encounters.

In video 2, the algorithm struggles because of the high number of customers and the fact that the customers are partly occluded
when they enter the shop. In addition, identity switches occur when the customer that has been served exits the shop
and in the process passes the other customers leading to the bounding boxes overlapping. An example of this can be seen in
Figure \ref{fig:objdetexample} where the customer at the entrance has not been detected by the algorithm. This leads to a
severe underestimation of the total number of customers 
and is reflected in the low recall for this particular video. Overall this points to two weaknesses of our methodology: in cases where
a high number of customers is present in the shop at the same time, the algorithm struggles to detect and track the customers
accurately, and in cases where an encounter takes a long time and the customer does not exit the shop immediately this leads to
a detection of the customer multiple times. In the next section of the paper, we discuss the implications of these limitations and
how to address them in future studies. 

Although the system is not designed nor optimized for real-time processing of video footage, the computational resources
used is still an important part of the evaluation of the system. As can be seen from Table \ref{table:runtime}, the system is able
to process on average 3 frames per second depending on the complexity of the footage. This means that e.g. video 2 with a length
of 30 minutes takes around 3 hours and 45 minutes to process. This is not problematic per se given the aim of the study which is
is to provide an estimate of the psychosocial work environment based on CCTV footage for use in research studies. In that respect,
runtime is not a critical factor for evaluating the usefulness of the system.

\begin{table}[h]
    \centering
    \caption{Runtime analysis of the system}
    \label{table:runtime}
    \renewcommand{\arraystretch}{1.5}
    \begin{tabular}{c c S[table-format=5.0] S[table-format=5.0] c}
    \toprule
    {Video Nr.} & {Length of video} & {\thead{No. of\\ frames}} & {\thead{Computational\\ time (secs)}} & {fps}  \\
    \midrule
    1 & 08:52 & 7993 & 2736 & 2.92 \\
    2 & 30:00 & 27002 & 13108 & 2.06  \\
    3 & 10:00 & 9002 & 1973 & 4.56 \\
    4 & 09:59 & 8999 & 2589 & 3.48 \\
    5 & 15:00 & 13502 & 6912 & 1.95 \\
    {Average} &  &  &  & 2.99 \\
    \bottomrule
    \end{tabular}
\end{table}

\subsection{Interactions between customers and employees}
\begin{figure}[h]
    \centering
    \includegraphics[width=0.49\textwidth]{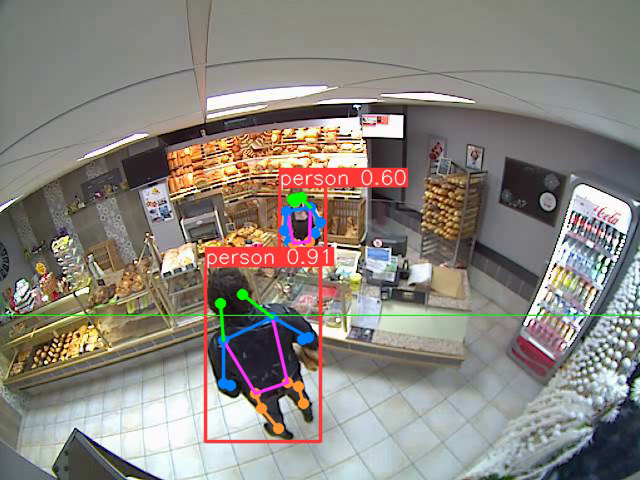}
    \caption{Example of pose estimation in the CCTV footage}
    \label{fig:poseexample}
\end{figure}

The results of the pose estimation algorithm and the rule-based approach to classify the interactions between customers and
employees are unfortunately not very promising at the current stage. While the algorithm is quite capable of estimating the poses of the customers,
the algorithm loses track of the employee in the frame at the current stage in a very high number of cases. The result shown in Figure \ref{fig:poseexample}
is therefore not representative of the actual results of the pose estimation algorithm where the employee falls in and out of the
frame in a high number of cases. This is likely due partly to the low quality of the footage and the fact that the employee is partly
occluded in the frame. In addition, the pose estimation algorithm is not able to estimate the poses of the persons in the frame
when they are too close to each other. This has the consequence that the estimated duration of the encounters based on the pose 
estimation severely underestimates the actual duration of the encounters in the footage. On average the estimated duration of the
encounters is 8 seconds while the actual duration of the encounters is 1 minute and 51 seconds. This is a major limitation of the 
methodology and more results of the pose estimation algorithm are therefore not included in the results of the study. Below we 
discuss whether this is due to a fundamental flaw in our proposed methodology or whether it is due to limitations of the
pose estimation algorithm and the quality of the footage used at present.

\section{Conclusion and Future Work}
This paper has contributed with a proof of concept for a methodology that can be used for estimating certain aspects of the
psychosocial work environment (more specifically quantitative job demands) using realistic CCTV footage. Drawing on several
existing computer vision algorithms, we propose a pipeline that detects and tracks persons in the footage, classifies them
as customers and employees respectively and based on these computations estimates core aspects of \(V_{\text{job demands}}\).
The results of the experiments carried out as part of this study show that the methodology is able to 
estimate the number of customers in the footage.

The results are best in videos with a
few costumers and a neutral interaction pattern. The algorithm struggles in cases with a high number of customers or
in situations where the duration of the encounters are longer than average. In particular, the algorithm fails in cases with
positive interactions
or when there are multiple customers waiting in line to be served. Then, at the current state, the pose estimation algorithm
is unable to distinguishing between positive, neutral and negative, 
likely due
to difficulties in estimating the pose of the employee in the frame. 

Therefore, even when the proposed solution
is promising, it cannot be used with `out-of-the-box',
so it still needs adjustments for the work place where it will be used,
where hand-crafted rules are used to classify the interactions between customers and employees, limiting its scalability. 

As future work, we suggest the following directions:

\begin{itemize}
    \item 
Combining footage from multiple cameras in a shop could improve detection accuracy, particularly in crowded settings where occlusions
are frequent, including more advanced tracking algorithms~\cite{omeragic_tracking_2020}

 \item 
Using a labeled dataset from the specific workplace to train models could improve interaction recognition. This includes annotating
poses indicative of various interaction types, thereby enhancing the method's ability to distinguish between positive, neutral, and
negative interactions, or employing deep learning models tailored to these specific interactions could yield better results~\cite{zheng_deep_2023}.

 \item 
Comparing the algorithm's output with self-reported data from employees can validate its effectiveness. Collecting employees' perceptions
of their job demands and interactions could provide a benchmark for refining the algorithm's accuracy. 

\end{itemize}

\bibliographystyle{ACM-Reference-Format} 
\bibliography{sw6} 

\end{document}